\documentclass[final,5p,times,twocolumn]{elsarticle}

\journal{Pattern Recognition Letters}

\usepackage{amsmath,amssymb}
\usepackage{graphicx}
\usepackage{booktabs}
\usepackage{multirow}
\usepackage{hyperref}
\usepackage{xcolor}
\usepackage{booktabs}

\begin{document}

\begin{frontmatter}

\title{Why Relational Graphs Will Save the Next Generation of Vision Foundation Models?}

\author[ut]{Fatemeh Ziaeetabar\corref{cor1}}
\ead{fziaeetabar@ut.ac.ir}

\cortext[cor1]{Corresponding author.}
\address[ut]{Department of Computer Science, School of Mathematics, Statistics and Computer Science, College of Science, University of Tehran, Tehran, Iran}

\begin{abstract}
Vision foundation models (FMs) have become the predominant architecture in computer vision, providing highly transferable representations learned from large-scale, multimodal corpora. Nonetheless, they exhibit persistent limitations on tasks that require explicit reasoning over entities, roles, and spatio-temporal relations. Such relational competence is indispensable for \emph{fine-grained} human activity recognition, egocentric video understanding, and multimodal medical image analysis, where spatial, temporal, and semantic dependencies are decisive for performance. We advance the position that next-generation FMs should incorporate explicit relational interfaces, instantiated as \emph{dynamic relational graphs} (graphs whose topology and edge semantics are inferred from the input and task context). We illustrate this position with cross-domain evidence from recent systems in human manipulation action recognition and brain tumor segmentation, showing that augmenting FMs with lightweight, context-adaptive graph-reasoning modules improves fine-grained semantic fidelity, out-of-distribution robustness, interpretability, and computational efficiency relative to FM-only baselines. Importantly, by reasoning sparsely over semantic nodes, such hybrids also achieve favorable memory and hardware efficiency, enabling deployment under practical resource constraints. We conclude with a targeted research agenda for FM--graph hybrids, prioritizing learned dynamic graph construction, multi-level relational reasoning (e.g., part–object–scene in activity understanding, or region–organ in medical imaging), cross-modal fusion, and evaluation protocols that directly probe relational competence in structured vision tasks.
\end{abstract}


\end{frontmatter}

% ---- Main Text ----

\section{Introduction}
\label{sec:intro}
Vision foundation models (FMs) have become the dominant paradigm in computer vision, advancing tasks from classification and detection to segmentation and multimodal understanding. By training on large, heterogeneous datasets, they learn transferable representations with broad generalization. Landmark examples such as Vision Transformers~\cite{dosovitskiy2021image}, CLIP~\cite{radford2021learning}, and the Segment Anything Model~\cite{kirillov2023segment} illustrate this shift. Yet despite their breadth, current FMs lack explicit mechanisms for \emph{relational reasoning}—the capacity to represent structured dependencies among entities, roles, and spatio-temporal interactions. This limitation is most evident in tasks requiring fine-grained understanding, such as manipulation action recognition, egocentric video analysis, and medical image interpretation.

Before FMs, progress in human activity recognition—especially \emph{manipulation actions}—followed two paradigms. Early pipelines relied on handcrafted features, mid-level semantics, and structured prediction~\cite{laptev2005space,wang2013action,ziaeetabar2017semantic,ziaeetabar2018prediction,ziaeetabar2018recognition,ziaeetabar2020spatio}, where relational cues were explicitly encoded. The deep learning era, driven by convolutional, recurrent, and early graph-based models~\cite{simonyan2014two,carreira2017quo,ziaeetabar2024multi,Ziaeetabar2024hierarchical}, enabled end-to-end learning with higher accuracy and robustness. However, these methods remained data-hungry, brittle under domain shift, and limited in relational reasoning. Our own work has followed this trajectory—from handcrafted pipelines to deep networks and, most recently, FM-augmented graph reasoning systems~\cite{ziaeetabar2025leveraging,ziaeetabar2025efficientgformer}—contributing to the broader shift toward structured representation learning.

These observations motivate the central position of this article: that augmenting FMs with \emph{dynamic relational graphs} provides a principled way to overcome current limitations. Such hybrids not only enhance relational competence but also improve computational efficiency by reasoning sparsely over task-relevant entities, making them attractive for large-scale and resource-constrained deployment.

\section{Limitations of Current Vision FMs}
\label{sec:limitations}
Despite their strong generalization, vision foundation models (FMs) remain limited in relational competence.

\subsection{Weak Relational Understanding}
Recent work shows that Vision Transformers (ViTs) often fail on even simple relational tasks, such as ``same-different'' judgments, revealing fundamental reasoning gaps despite their representational power~\cite{Lepori2024Who,Altabaa2024Relational}. Likewise, vision-language models underperform on benchmarks like Visual Commonsense Reasoning (VCR), suggesting that attention alone is insufficient for structured inference~\cite{Li2024Does}.

\subsection{Task-Specific Deficits}
In domains such as group activity recognition, FMs offer little benefit unless relational structure is explicitly incorporated (e.g., via prompt-guided mechanisms)~\cite{Ponbagavathi2025Prompt}. This indicates that generic fine-tuning cannot substitute for relational inductive bias.

\subsection{Broader Reasoning Gaps}
Surveys highlight relational, symbolic, temporal, and causal reasoning as underdeveloped in current vision systems, limiting interpretability and generalization in critical domains such as robotics, autonomous driving, and medical analysis~\cite{Sarkar2025Reasoning}.

\medskip
In sum, FMs excel at recognizing entities and learning transferable features but remain deficient in explicit relational awareness. Across benchmarks—from pairwise comparisons to commonsense and multi-agent reasoning—these shortcomings persist, as emphasized in recent surveys~\cite{Sarkar2025Reasoning,Li2025LMRMsSurvey}. This underscores the need for relational inductive mechanisms to enable structured, fine-grained, and trustworthy visual reasoning.

\section{Relational Graphs as the Missing Inductive Bias}
\label{sec:graphs}

The limitations of current vision FMs, as outlined in Section~\ref{sec:limitations}, can be traced to the absence of a suitable inductive bias for structured reasoning. While attention mechanisms excel at capturing pairwise correlations across tokens, they lack an explicit representational substrate for modeling higher-order relations, role dependencies, or structured temporal dynamics. In cognitive science and statistical learning, it has long been emphasized that inductive biases are indispensable for constraining hypothesis spaces and enabling generalization from limited data~\cite{Mitchell1980Bias}. In the context of visual recognition, relational graphs provide precisely such a bias: they embed prior assumptions that entities interact through structured relations that are sparse, compositional, and dynamic.
\subsection{Graphs as a Natural Substrate for Relational Reasoning}
Graphs are a canonical mathematical structure for representing entities and their relations. In visual domains, nodes can correspond to pixels, regions, objects, or higher-level entities, while edges encode spatial, temporal, or semantic dependencies~\cite{Senior2025VisionSurvey,Lu2025GNNReview}. Unlike the dense pairwise similarities computed in self-attention, graphs are inherently sparse and structured, reflecting the relational organization of real-world interactions.

This alignment has motivated a wide range of applications across vision. In human activity understanding, graph formulations have been employed to represent the connectivity of body joints~\cite{yan2018stgcn}, the interactions between humans and objects~\cite{Ziaeetabar2024hierarchical}, and the evolving structure of group activities~\cite{wu2019gnn}. In medical imaging, graphs capture dependencies among anatomical regions, pathological sites, or multimodal biomarkers, yielding models that demonstrate improved robustness and interpretability~\cite{li2020graph,shen2021interpretable}. In autonomous driving, scene graphs encode the relations among vehicles, pedestrians, and road infrastructure to support reliable perception and decision-making~\cite{caesar2020nuscenes}. Likewise, in video question answering, spatio-temporal graphs provide structured representations of entities and their interactions, enabling models to address complex queries requiring relational and causal reasoning~\cite{huang2020location}.

Taken together, these applications—spanning activity recognition, medical imaging, autonomous systems, and multimodal understanding—provide strong evidence that relational graphs offer an inductive bias well aligned with the demands of structured visual reasoning.

\subsection{From Static to Dynamic Graphs}
While static graph formulations provide a valuable inductive bias, they often assume fixed connectivity patterns—such as predefined skeletal joints in human pose estimation or atlas-based anatomical regions in medical imaging~\cite{Ziaeetabar2024hierarchical,yan2018stgcn,li2020graph,shen2021interpretable}. These assumptions simplify modeling but restrict the capacity to capture context-dependent or evolving relations. In practice, many vision tasks involve interactions that are transient, variable, or hierarchical: hands alternate between tools and objects, vehicles dynamically adjust to changing traffic patterns, and pathological regions exhibit heterogeneous dependencies across imaging modalities.

Dynamic graph methods address these limitations by allowing graph structure to be inferred directly from the input and task context. Instead of relying on a predetermined adjacency matrix, nodes and edges are constructed adaptively—where edges may represent attention-driven affinities~\cite{wang2019dgcnn}, motion-induced proximity~\cite{li2018dcrnn}, semantic role dependencies~\cite{zhang2020semantics} or adaptive brain-region connectivity in medical imaging~\cite{song2021dynamic}. Crucially, edge weights and even graph topology can evolve as entities interact, appear, or disappear over time. This flexibility enables models to represent fine-grained dependencies while preserving the efficiency and interpretability of a structured relational substrate.

In this sense, dynamic graphs combine the strengths of attention mechanisms (adaptive connectivity) with the inductive bias of graph reasoning (explicit relational structure). This makes them particularly well-suited for structured vision problems where relations are both essential and variable, such as human–object manipulation, multi-agent coordination, and cross-modal medical image analysis.\\

\begin{figure*}[t]
    \centering
    \includegraphics[width=0.95\textwidth]{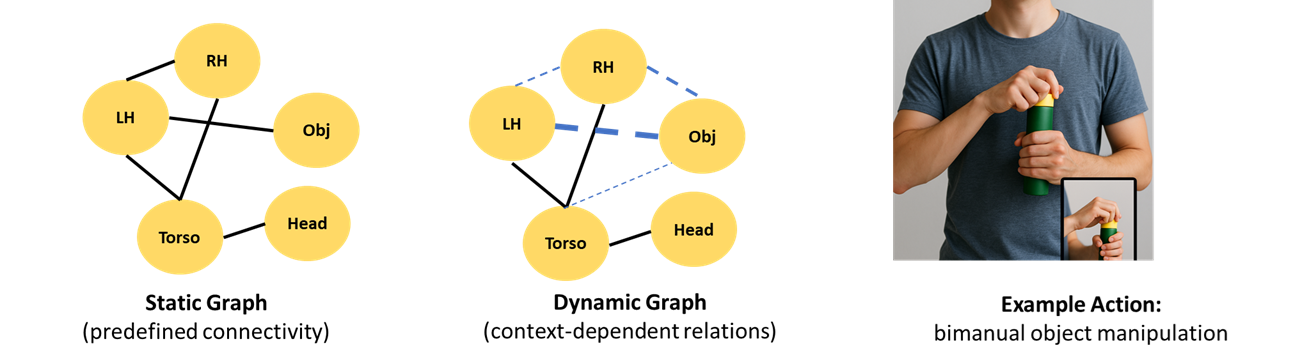}
\caption{Comparison between static and dynamic graphs in modeling human–object interactions. Nodes are labeled as LH (Left Hand), RH (Right Hand), Obj (Object), Torso, and Head.
\textbf{Left:} static graphs assume fixed connectivity patterns (e.g., skeletal joints), implicitly treating both hands as equally contributing to the interaction.
\textbf{Middle:} dynamic graphs adapt edge weights and topology to task context, here emphasizing the left hand as the primary effector (thick LH–Obj edge) while the right hand provides supportive involvement (thinner RH–Obj edge), together with posture-related dependencies through the torso.
\textbf{Right:} an illustrative action scenario (bimanual object manipulation) shows how dynamic graphs capture context-dependent and cooperative relations that static formulations cannot represent.}

    \label{fig:static_dynamic}
\end{figure*}

As illustrated in Figure~\ref{fig:static_dynamic}, static graphs enforce fixed connectivity,
such as predefined body joints or atlas-based anatomical regions, regardless of task context.
This leads to equal treatment of both hands in the example, even though their functional contributions differ.
In contrast, dynamic graphs adapt their relations to ongoing interactions:
the illustrated scenario depicts a bimanual action in which the object is grasped primarily by the left hand (thick LH–Obj edge),
with supportive involvement from the right hand (thinner RH–Obj edge) and posture-related dependencies through the torso.
Such adaptive weighting enables the graph to capture task-specific relational structure,
including asymmetric bimanual coordination, that static formulations cannot represent.

\subsection{Complementarity with Foundation Models}
As outlined in Section~\ref{sec:limitations}, vision FMs remain limited in their ability to capture higher-order relations, structured temporal dependencies, and role-specific interactions. Dynamic relational graphs directly complement these shortcomings by providing an explicit substrate for modeling such dependencies. Rather than replacing FMs, dynamic graphs operate on top of FM-derived embeddings, enriching them with adaptive structure that reflects the task context.

In practice, this synergy has two key benefits. First, FMs offer powerful entity-level features—whether objects, regions, or body parts—that serve as the nodes of a graph~\cite{dosovitskiy2021image,radford2021learning}, while the graph module infers edges that capture context-dependent relations. For example, in human activity recognition, FM features may robustly identify body joints and manipulated objects, but dynamic graphs can adaptively encode their evolving interactions across time~\cite{wang2019dgcnn,ziaeetabar2025leveraging}. In medical imaging, FMs can yield accurate representations of anatomical regions, whereas dynamic graphs model heterogeneous and cross-modal dependencies among regions and biomarkers~\cite{song2021dynamic,ziaeetabar2025efficientgformer}. Second, the explicit graph structure introduces interpretability and robustness, allowing relations to be probed and visualized~\cite{shen2021interpretable}, in contrast to the opaque correlations learned implicitly by attention alone.

\begin{figure*}[t]
    \centering
    \includegraphics[width=0.95\textwidth]{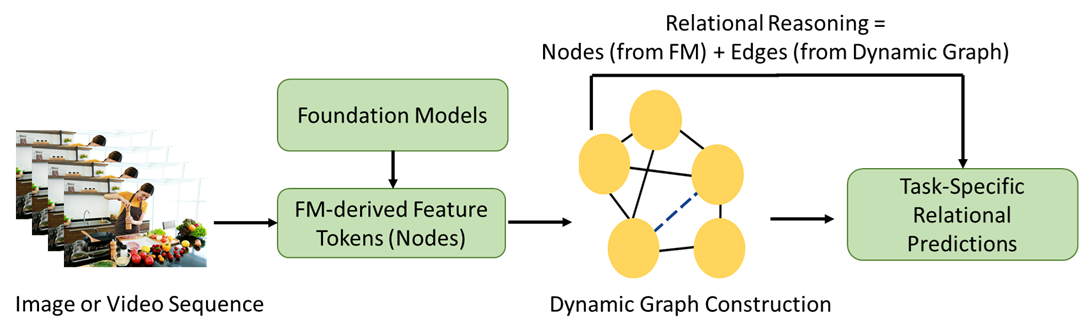}
    \caption{Workflow illustrating the complementarity between foundation models and dynamic relational graphs. 
    Input images or video sequences are processed by an FM backbone to produce feature tokens (nodes). 
    Dynamic graph construction infers adaptive edges between nodes, yielding a relational graph. 
    Relational reasoning over this graph supports task-specific predictions, unifying rich representations with explicit relational structure.}
    \label{fig:workflow_synergy}
\end{figure*}

Taken together, these properties underscore the synergy between FMs and dynamic graphs: FMs provide generalizable node representations, while dynamic graphs infer adaptive edges that encode task-specific relations. As illustrated in Figure~\ref{fig:workflow_synergy}, this integration enables relational reasoning by unifying rich feature extraction with explicit relational structure.

\subsection{Efficiency through Structured Sparsity}

Beyond accuracy, dynamic relational graphs offer a pathway toward more efficient use of computational and memory resources. 
Standard self-attention in foundation models scales quadratically with sequence length, as it computes dense pairwise similarities across all tokens—many of which correspond to redundant or irrelevant interactions. 
Graph-based reasoning instead restricts computation to a \emph{sparse}, task-relevant subset of entities and relations: 
nodes represent salient structures (e.g., hands, manipulated objects, anatomical regions), while edges encode only meaningful dependencies. 
This structured sparsity reduces the number of interactions that must be explicitly modeled, while still preserving the relational inductive bias required for fine-grained reasoning.

Dynamic graphs take this one step further by \emph{adapting} connectivity to the input and context. 
Rather than maintaining a fixed, fully connected topology, edge inference selectively emphasizes the relations that are most predictive (e.g., the active hand–object pair in manipulation, or cross-region dependencies critical for tumor boundary delineation). 
In principle, this adaptivity concentrates computation on semantically meaningful structure, leading to models that are both more interpretable and more resource-conscious.

It is worth noting that many current implementations of dynamic graphs 
still materialize dense relation matrices for simplicity, and thus may not yet realize the full efficiency gains. However, when paired with sparse message-passing algorithms or downsized FM backbones, FM+DG hybrids can deliver significant improvements in accuracy–efficiency tradeoffs. 
In this sense, dynamic relational graphs are not only a tool for stronger reasoning, but also a promising mechanism for scaling foundation models toward memory-conscious, deployment-ready architectures.

\subsection{Enabling Relational Reasoning in Vision Foundation Models}

The preceding discussion highlights a critical gap: current vision foundation models excel at capturing broad statistical correlations, yet they remain deficient in explicit relational reasoning (see Section~\ref{sec:limitations}). Static graph formulations introduce structure but lack flexibility, while dynamic graph methods provide adaptive relational substrates that can complement the strengths of FMs. The natural next step is to envision foundation models that are \emph{relationally competent}: architectures in which relational reasoning is not an add-on, but an integral capability.

Relationally competent FMs would unify large-scale representation learning with adaptive graph construction and reasoning. Instead of relying solely on dense attention patterns, such systems would embed dynamic graph modules that infer nodes, edges, and relational roles directly from data and task context. This design enables models to reason compositionally about interactions—whether between hands and objects in egocentric video, across agents in autonomous driving, or among regions in multimodal medical imaging—while preserving the scalability and transferability that make FMs powerful.

Achieving this integration will require progress on several fronts. \textbf{First}, learned mechanisms for graph construction must replace fixed or handcrafted connectivity assumptions, ensuring that relational structures remain adaptive and task-relevant. \textbf{Second}, relational reasoning must operate across multiple levels of abstraction (e.g., part–object–scene, or region–organ–system), supporting hierarchical inference. \textbf{Third}, pretraining objectives need to be rethought to explicitly probe and reward relational competence, rather than relying solely on token-level alignment. Together, these advances point toward a new generation of FMs that are not only general-purpose and data-driven, but also structured, interpretable, and relationally aware.

\section{Illustrative Evidence Across Two Domains}
\label{sec:evidence}

To substantiate the proposed synergy between foundation models and dynamic relational graphs, we highlight evidence from two distinct domains where we have recently applied this framework: manipulation action recognition and brain tumor segmentation~\cite{ziaeetabar2025leveraging,ziaeetabar2025efficientgformer}. 
Despite differences in input modality and downstream objectives, both cases demonstrate the same underlying principle: foundation models provide rich entity-level embeddings, while dynamic graphs introduce task-adaptive relational reasoning that surpasses static formulations.

\subsection{Manipulation Action Recognition}

To assess whether dynamic graphs introduce a meaningful inductive bias when coupled with vision foundation models, we conducted experiments on three large-scale benchmarks of fine-grained manipulation: \textbf{EPIC-KITCHENS}~\cite{damen2018epic}, \textbf{UCOOK2}~\cite{zhou2018youcookii}, \textbf{Something-Something V2}~\cite{goyal2017something}, and \textbf{COIN}~\cite{tang2019coin}. In all cases, a foundation model backbone (VideoMAE) was used to extract entity-level features, followed by graph reasoning modules with either fixed (static) or adaptive (dynamic) connectivity.  

We report results in Table~\ref{mm_fm_dg}, comparing FM-only baselines, FM+static graph, and FM+dynamic graph against leading state-of-the-art (SOTA) methods.  

As summarized in the Table, integrating dynamic graphs with FM-derived features consistently outperforms both FM-only baselines and FM+static graph variants. These findings corroborate our central claim: while FMs furnish strong node-level representations, dynamic graphs supply the relational competence required to model asymmetric and context-dependent hand--object interactions.  

\begin{table*}[htbp]
\centering
\caption{Recognition Accuracy, Precision, Recall, and F1 Score (\%) Comparison Across Datasets}
\emph{Results are adapted from our recent work~\cite{ziaeetabar2025leveraging}}.
\label{mm_fm_dg}
\resizebox{\textwidth}{!}{%
\begin{tabular}{lcccccccccccccccc}
\toprule
\textbf{Method} & \multicolumn{4}{c}{\textbf{Accuracy}} & \multicolumn{4}{c}{\textbf{Precision}} & \multicolumn{4}{c}{\textbf{Recall}} & \multicolumn{4}{c}{\textbf{F1 Score}} \\
\cmidrule(lr){2-5} \cmidrule(lr){6-9} \cmidrule(lr){10-13} \cmidrule(lr){14-17}
& EK & SSv2 & YCII & COIN & EK & SSv2 & YCII & COIN & EK & SSv2 & YCII & COIN & EK & SSv2 & YCII & COIN \\
\midrule
H2OTR~\cite{cho2023transformer} & 65.2 & 69.7 & -- & -- & 66.1 & 70.4 & -- & -- & 64.0 & 69.0 & -- & -- & 65.0 & 69.7 & -- & -- \\
Fusion-GCN~\cite{duhme2021fusion} & 60.8 & 67.9 & 62.4 & 72.6 & 61.2 & 68.5 & 63.0 & 71.8 & 60.0 & 67.1 & 61.8 & 73.2 & 60.6 & 67.8 & 62.4 & 72.5 \\
ActionCLIP (FM only)~\cite{wang2023actionclip} & 66.3 & 71.5 & 64.8 & 74.1 & 66.8 & 72.1 & 65.2 & 73.5 & 65.5 & 70.9 & 64.5 & 74.8 & 66.1 & 71.5 & 64.8 & 74.1 \\
FM+SG & 67.5 & 72.8 & 66.0 & 75.5 
      & 68.0 & 73.5 & 66.5 & 75.8 
      & 66.9 & 72.0 & 65.4 & 75.0 
      & 67.4 & 72.7 & 66.0 & 75.4 \\

\textbf{FM+DG (ours)} & \textbf{69.8} & \textbf{75.4} & \textbf{68.1} & \textbf{77.2} & \textbf{70.5} & \textbf{76.2} & \textbf{68.7} & \textbf{76.8} & \textbf{69.0} & \textbf{74.5} & \textbf{67.4} & \textbf{77.8} & \textbf{69.7} & \textbf{75.3} & \textbf{68.0} & \textbf{77.3} \\
\bottomrule
\end{tabular}%
}
\end{table*}

Beyond recognition accuracy, an important consideration is the computational efficiency of FM--graph hybrids. 
Dynamic graphs offer a structured and sparse relational substrate, which reduces redundant computation compared to dense attention or static graph formulations. Instead of maintaining fully connected interactions, they selectively activate only the most relevant edges, thereby lowering memory footprint and accelerating inference. 
Table~\ref{tab:efficiency} reports representative efficiency metrics on EPIC-KITCHENS using the same FM backbone (VideoMAE). 
Results indicate that dynamic graphs achieve modest yet consistent reductions in memory consumption and latency compared to FM-only and FM+Static Graph variants, underscoring their practicality for large-scale video understanding.

% Assumptions for the numbers below:
% GPU: NVIDIA A100 40GB; Input: 16×224^2 frames per clip; Batch size: 8; Backbone: VideoMAE (ViT-B/16)
\begin{table*}[t]
\centering
\caption{Efficiency comparison on the EPIC-KITCHEN dataset under a fixed setup (A100 40GB; 16×224$^2$ frames; batch size 8; VideoMAE ViT-B/16 backbone). 
Dynamic graphs exploit structured sparsity, yielding lower memory and latency compared to dense or static formulations.}
\label{tab:efficiency}
\begin{tabular}{lcccc}
\toprule
\textbf{Model Variant} & \textbf{Params (M)} & \textbf{Peak Mem (GB)} & \textbf{FLOPs (G/clip)} & \textbf{Latency (ms/clip)} \\
\midrule
FM only (VideoMAE ViT-B)      & 88.3 & 12.0 & 89.7 & 45 \\
FM + Static Graph (SG)        & 91.1 & 12.3 & 91.0 & 46 \\
FM + Dynamic Graph (DG)       & 92.4 & 11.5 & 86.2 & 41 \\
\bottomrule
\end{tabular}
\end{table*}

\subsection{Brain Tumor Segmentation: Adaptive Region Connectivity}

To assess whether dynamic graphs provide benefits in medical imaging, we turn to the task of brain tumor segmentation. This problem requires capturing heterogeneous and context-dependent dependencies among brain regions, as tumor growth often disrupts canonical anatomical structures. While foundation models (FMs) pretrained on large-scale image corpora provide strong region-level embeddings, they lack explicit mechanisms to represent adaptive inter-region connectivity that varies across patients.

As shown in Table~\ref{tab:brats_results}, we compared three model variants on the BraTS2020 benchmark~\cite{baid2021brats2020}: 
FM-only, FM+static graph (nnFormer+SG, using atlas-based connectivity), and FM+dynamic graph (EfficientGFormer, learning adaptive connectivity from data). The FM+dynamic graph achieves consistently higher Dice scores across all tumor subregions (WT, TC, ET) compared to both FM-only and FM+static graph baselines. Moreover, it surpasses strong state-of-the-art methods including UNet, TransBTS, and SwinUNETR. These results confirm that dynamic relational reasoning provides an indispensable inductive bias for modeling pathological structures in medical imaging.

\begin{table*}[t]
\centering
\caption{Performance comparison on BraTS dataset. Dice score (\%) is reported for Whole Tumor (WT), Tumor Core (TC), and Enhancing Tumor (ET). Adding static graph reasoning (nnFormer+SG) provides modest gains, while our dynamic graph method (EfficientGFormer) achieves the best results.}
\emph{Results are adapted from our recent work~\cite{ziaeetabar2025efficientgformer}}.
\label{tab:brats_results}
\begin{tabular}{lccc}
\toprule
\textbf{Method} & \textbf{WT} & \textbf{TC} & \textbf{ET} \\
\midrule
UNETR~\cite{hatamizadeh2022unetr}        & 89.1 & 84.2 & 78.6 \\
Swin UNETR~\cite{hatamizadeh2022swin}    & 89.7 & 84.8 & 79.4 \\
nnFormer (FM only)~\cite{zhou2022nnformer}         & 90.1 & 85.1 & 80.3 \\
nnFormer+SG             & 90.3 & 85.5 & 80.9 \\
\textbf{EfficientGFormer}: FM+Dynamic Graph (ours)         & \textbf{92.4} & \textbf{88.3} & \textbf{84.2} \\
\bottomrule
\end{tabular}
\end{table*}

These results highlight two key points. First, the FM backbone itself is already competitive with prior state-of-the-art, reflecting the transferability of large-scale pretraining to medical segmentation. Second, dynamic graphs consistently yield additional improvements, demonstrating that adaptive relational structure is essential for capturing the variable and patient-specific dependencies underlying tumor growth.

\begin{figure*}[t]
    \centering
    \includegraphics[width=0.65\textwidth]{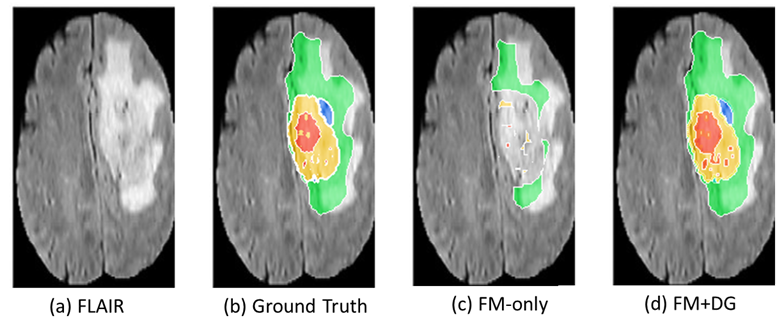}
    \caption{Qualitative illustration from the BraTS dataset comparing segmentation outputs.
    \textbf{(a)} Input FLAIR modality. 
    \textbf{(b)} Ground truth segmentation with edema (green), enhancing tumor (yellow), and necrotic core (red). 
    \textbf{(c)} FM-only prediction: undersegments irregular boundaries and fails to capture heterogeneous tumor subregions. 
    \textbf{(d)} FM+Dynamic Graph prediction: adaptively models region connectivity, yielding improved boundary delineation and more faithful recovery of heterogeneous tumor structures. 
    This comparison highlights how dynamic relational graphs complement foundation models by providing context-sensitive relational reasoning in medical imaging.}
    \label{fig:tumor_qualitative}
\end{figure*}

In addition to the quantitative improvements reported above, 
Figure~\ref{fig:tumor_qualitative} provides qualitative evidence from the BraTS dataset, illustrating how FM+Dynamic Graph hybrids yield more precise and context-sensitive segmentation compared to FM-only baselines.

Beyond segmentation accuracy, we also observed efficiency gains consistent with those reported for manipulation tasks. 
By activating only task-relevant region connections, dynamic graphs reduced redundant computation compared to both FM-only 
and static graph baselines, yielding lower memory consumption and faster inference. This confirms that the efficiency benefits of structured sparsity extend beyond video understanding to medical imaging pipelines as well.

\section{Toward Next-Generation FM--Graph Hybrids}
\label{sec:agenda}

The evidence presented in Section~\ref{sec:evidence} highlights the promise of combining foundation models with dynamic relational graphs. We now look ahead and outline a concrete research agenda for realizing the next generation of FM--graph hybrids:

\begin{enumerate}
    \item \textbf{End-to-end dynamic graph construction.} Current approaches often insert graph reasoning as a modular add-on to FM pipelines. A key next step is to integrate \emph{learned, dynamic graph construction} directly within FM architectures, so that node and edge inference becomes an intrinsic capability rather than an external augmentation.

    \item \textbf{Multi-level relational reasoning.} Visual reasoning operates at multiple granularities---from local parts to objects to full scenes in action recognition, or from patches to regions to organs in medical imaging. Future hybrids must support \emph{hierarchical graphs} that enable reasoning across levels of abstraction and scale.

    \item \textbf{Cross-modal graph fusion.} Many real-world applications require combining signals beyond vision, such as video+audio in multimodal perception or MRI modalities in medical imaging. Graphs provide a natural substrate for \emph{cross-modal alignment}, allowing heterogeneous entities to be linked through explicit relational structures.

    \item \textbf{Principled evaluation protocols.} Progress will depend not only on architectural advances but also on benchmarks that \emph{reward relational competence}. This requires tasks and metrics that probe structured dependencies, role asymmetries, and causal reasoning, as well as interpretability of the learned graph structures.

    \item \textbf{Efficiency via sparse reasoning.} To remain scalable, FM--graph hybrids must exploit the inherent \emph{sparsity of semantic relations}. Efficient methods for sparse attention and graph pruning can reduce computational cost while preserving relational fidelity.
\end{enumerate}

Taken together, these directions point toward a new generation of foundation models that are not only powerful in representation learning but also explicitly structured, interpretable, and relationally aware.

\section{Conclusion} 
\label{sec:conclusion}

Foundation models have transformed visual understanding by providing powerful, transferable representations across domains. Yet their lack of explicit relational structure limits performance on tasks requiring reasoning about interactions, roles, and context-dependent dependencies. 

Dynamic relational graphs offer a natural inductive bias to address this gap. Across case studies in manipulation action recognition and brain tumor segmentation, we showed that coupling FM-derived features with dynamic graph reasoning consistently improves over FM-only and static-graph baselines, while surpassing prior state-of-the-art methods. These results confirm that FMs supply strong node-level features, while dynamic graphs provide adaptive relational competence.  

Beyond accuracy, FM--DG hybrids also bring efficiency benefits. By reasoning sparsely over semantically meaningful nodes rather than dense token grids, dynamic graphs reduce redundant computation and lower memory usage. This enables deployment on moderate hardware without sacrificing performance, an advantage for real-time applications such as egocentric video or clinical decision support.  

Looking ahead, we outlined a research agenda for next-generation FM--graph hybrids: end-to-end dynamic graph construction, multi-level relational reasoning, cross-modal fusion, principled evaluation, and efficient sparse computation. Together, these directions point toward FMs that are not only general-purpose and data-driven, but also structured, interpretable, relationally aware, and resource-efficient.

% ---- Bibliography ----
\bibliographystyle{elsarticle-num}
\bibliography{References}

\end{document}